%% file: main.tex
\documentclass[10pt,twocolumn,letterpaper]{article}

\usepackage[applications]{wacv}      
\usepackage{xcolor,colortbl}
\usepackage[accsupp]{axessibility}  

\input{preamble}

\definecolor{wacvblue}{rgb}{0.21,0.49,0.74}
\usepackage[pagebackref,breaklinks,colorlinks,allcolors=wacvblue]{hyperref}

\def\wacvPaperID{7} 
\def\confName{WACV}
\def\confYear{2026}

\title{When Humans Judge Irises:\\Pupil Size Normalization as an Aid and Synthetic Irises as a Challenge}

\author{Mahsa Mitcheff and Adam Czajka\\
384 Fitzpatrick Hall of Engineering, University of Notre Dame, IN 46556, USA\\
{\tt\small \{mmitchef,aczajka\}@nd.edu}
}

\begin{document}
\maketitle
\input{sec/0_abstract}    
\input{sec/1_intro}
\input{sec/2_related_work}
\input{sec/3_experiments}
\input{sec/4_dataset}

\input{sec/5_results}
\input{sec/6_conclusion}
{
    \small
    \bibliographystyle{IEEEtran}
    \bibliography{main}
}

\end{document}


\maketitle

This document presents the materials supplementing the results and illustrations provided in the main paper.

\begin{figure}[!htbp]
    \centering
\includegraphics[width=1.0\linewidth]{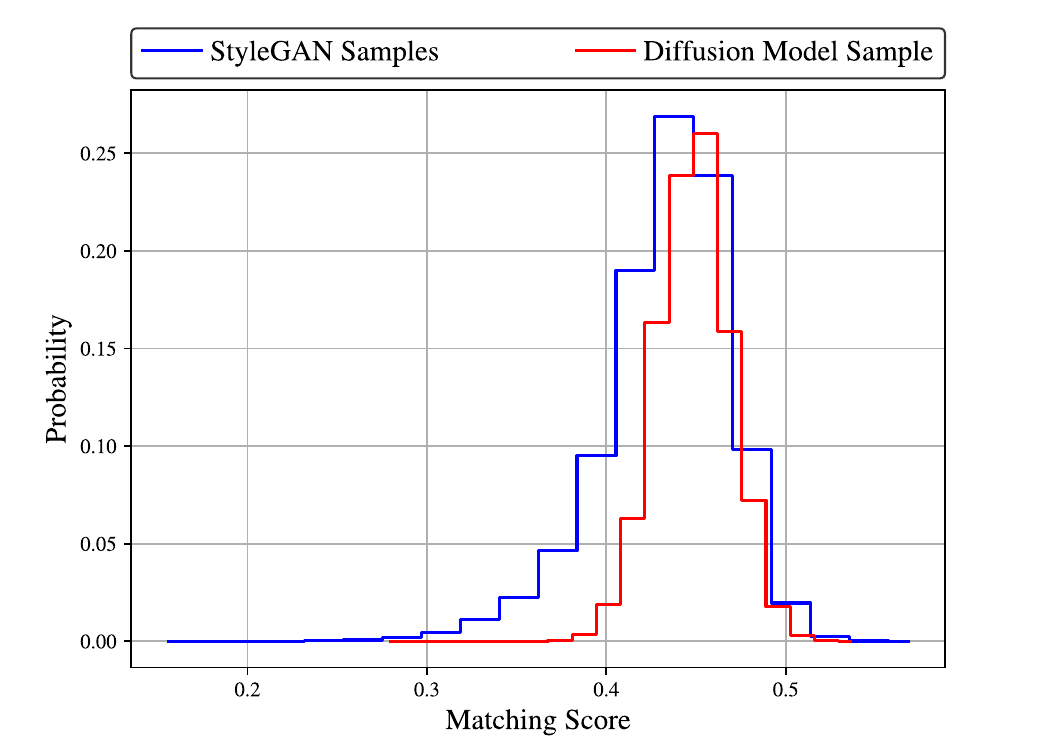}
    \caption{HDBIF scores for synthetic iris samples generated by StyleGAN and Diffusion models.}
    \label{fig:GAN_DM_Score}
\end{figure}

Figure~\ref{fig:GAN_DM_Score} illustrates the distribution of matching scores computed using the HDBIF method for 5,000 synthetic samples generated by StyleGAN and 5,000 samples produced by the diffusion model. As observed in the figure, the StyleGAN model generates synthetic samples with a lower matching score compared to those generated by the diffusion model. A lower HDBIF matching score is interpreted as an indicator of higher authenticity.

\begin{figure*}[!htbp]
    \centering
\includegraphics[width=1.0\linewidth]{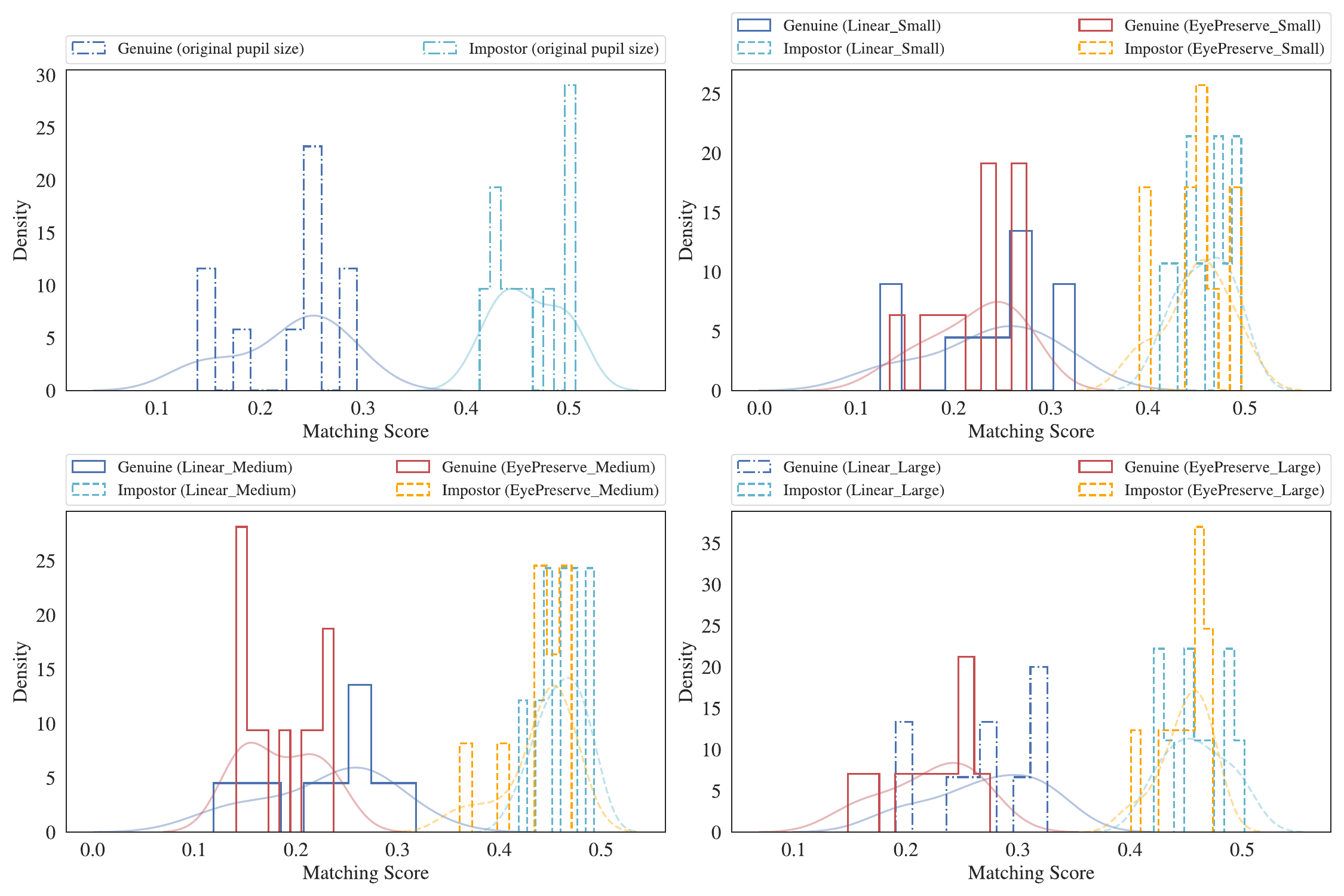}
    \caption{HDBIF scores for the selected samples in Scenario I.}
    \label{fig:pdm_score_matching_all_curves}
\end{figure*}

Figure~\ref{fig:pdm_score_matching_all_curves} presents the matching scores computed using the HDBIF method for selected irises in Scenario I across all six sample groups which includes irises from both genuine and impostor samples deformed by either EyePreserve or linear models under three different pupil sizes: \say{small}, \say{middle}, \say{large}. The resulting distributions illustrate the degree of separation between the selected genuine and impostor samples. Genuine samples deformed by the non-linear model yielded lower matching scores than those deformed by the linear model. In contrast, for impostor samples with small or medium pupil sizes, the non-linear model obtained higher matching scores.

\begin{figure}[!htbp]
    \centering
    \includegraphics[width=1.0\linewidth]{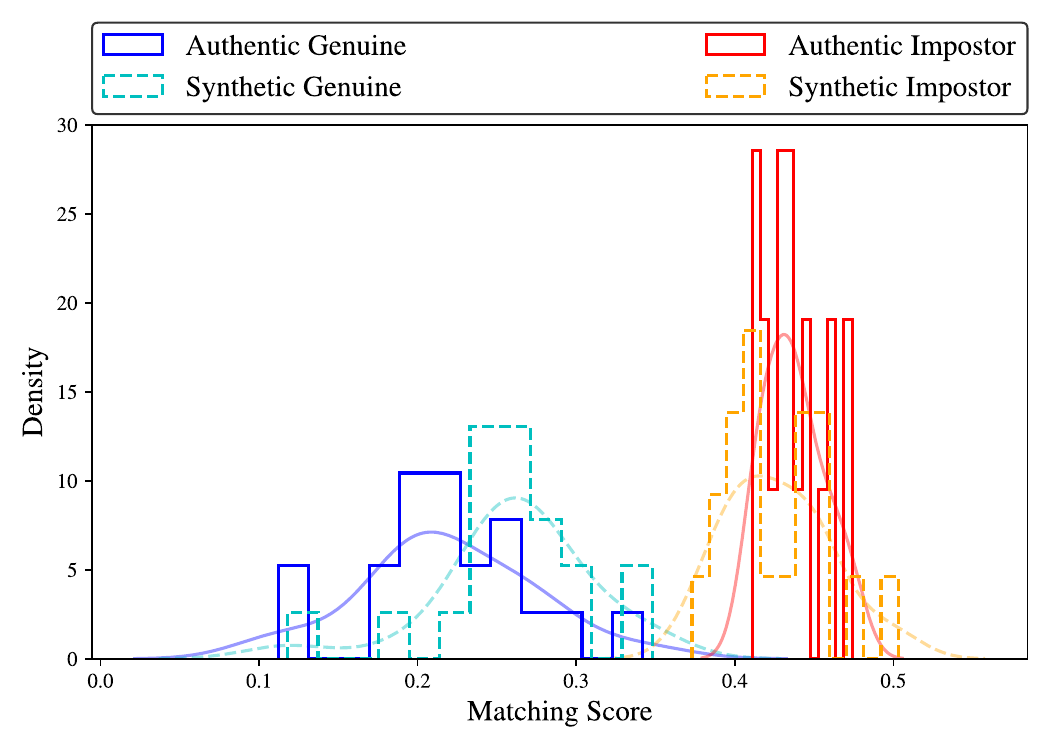}
    \caption{HDBIF scores for the selected samples in Scenario II.}
    \label{fig:iis_score_matching_all_curves}
\end{figure}

Figure~\ref{fig:iis_score_matching_all_curves} presents the matching scores computed using the HDBIF method for Scenario II across all four sample groups: \say{authentic-genuine}, \say{authentic-impostor}, \say{synthetic-genuine}, and \say{synthetic-impostor}. Most synthetic–genuine sample pairs showed matching scores around $0.3$, while impostor samples showed scores closer to $0.5$. The resulting distributions illustrate the degree of separation between the selected genuine and impostor samples.

%% file: preamble.tex
\usepackage{graphicx}
\usepackage{subcaption}
\usepackage{threeparttable}
\usepackage{multirow}
\usepackage{multicol}
\usepackage{comment}
\usepackage{makecell}
\usepackage{enumitem}
\usepackage{MnSymbol}
\usepackage{url}
\usepackage{booktabs}
\usepackage{svg}
\usepackage{dirtytalk}
\usepackage{csquotes}
\usepackage{xspace}
\usepackage{acro}
\usepackage[table]{xcolor}
\usepackage[accsupp]{axessibility}  


%% file: sec/0_abstract.tex
\begin{abstract}
Iris recognition is a mature biometric technology offering remarkable precision and speed, and allowing for large-scale deployments to populations exceeding a billion enrolled users (e.g., AADHAAR in India). However, in forensic applications, a human expert may be needed to review and confirm a positive identification before an iris matching result can be presented as evidence in court, especially in cases where processed samples are degraded (e.g., in post-mortem cases) or where there is a need to judge whether the sample is authentic, rather than a result of a presentation attack. 

This paper presents a study that examines human performance in iris verification in two controlled scenarios: (a) under varying pupil sizes, with and without a linear/nonlinear alignment of the pupil size between compared images, and (b) when both genuine and impostor iris image pairs are synthetically generated. The results demonstrate that pupil size normalization carried out by a modern autoencoder-based identity-preserving image-to-image translation model significantly improves verification accuracy. Participants were also able to determine whether iris pairs corresponded to the same or different eyes when both images were either authentic or synthetic. However, accuracy declined when subjects were comparing authentic irises against high-quality, same-eye synthetic counterparts. These findings (a) demonstrate the importance of pupil-size alignment for iris matching tasks in which humans are involved, and (b) indicate that despite the high fidelity of modern generative models, same-eye synthetic iris images are more often judged by humans as different-eye images, compared to same-eye authentic image pairs. 

We offer data and human judgments along with this paper to allow full replicability of this study and future works.
\end{abstract}

%% file: sec/1_intro.tex
\section{Introduction}
\label{sec:intro}
\subsection{Motivation}

Forensic iris analysis, which is the comparative analysis of pairs of irises to determine if they originate from (1) the same individual, (2) different individuals, or (3) images of insufficient quality to support a conclusion, serves as a highly established and superior forensic identification tool compared to many other biometric modalities ~\cite{matey2022forensic,bhatt2025systematic}. This comparison task can be performed using automated iris biometric systems, human examiners, or a combination of both~\cite{matey2022forensic}. However, in a court setting, there may be a need for a human expert to make the final decision on a positive identification (potentially supported by an automated system). Hence, despite an unquestionable success of automated iris recognition systems, there is a need for human iris recognition experts and studies that assess the experts' reliability under real-world complications, such as processing AI-generated samples, or images representing typical yet severe biology-driven deformations, such as pupil size variation.

Prior studies suggest that human subjects possess complementary discriminatory capabilities when processing iris images. For instance, research has shown that automated iris systems treat an individual's left and right irises as distinct identities, whereas human observers can often perceive similarities in the texture ignored by algorithms~\cite{bowyer2010human}. Furthermore, Hollingsworth et al.~\cite{hollingsworth2011genetically} demonstrated that even when automated systems classify the irises of identical twins as non-matches, which is certainly a correct outcome, humans are capable of discerning subtle iris texture patterns shared by related eyes. These findings suggest that human examiners utilize iris texture clues or a broader perceptual mechanism that algorithms ignore. 

However, the existing studies assessing human performance in iris recognition are rather sparse, and limited to just addressing selected problems, \eg listing categories of iris deformations posing the biggest challenges to humans, such as variable pupil size \cite{moreira2019performance}.
\subsection{Study Design and Contributions}

This paper offers two novel studies investigating human performance in iris recognition in two scenarios involving modern Artificial Intelligence (AI):

\begin{itemize}
    \item {\bf Scenario I}, described in Sec. \ref{Exp1-PD}, in which we mitigate the challenges associated with variable pupil size in images compared by experts by pupil size alignment utilizing linear deformation and image-to-image translation, identity-preserving nonlinear model of iris texture deformation, and
    \item {\bf Scenario II}, described in Sec. \ref{Exp1-AS}, in which we assess the human performance in comparing same-eye and different-eye AI-generated iris images, and comparing such performance with accuracy achieved for authentic iris image pairs (again: same-eye and different-eye ones).
\end{itemize}

\noindent To the best of the authors' knowledge, the above studies have not yet been conducted. Especially, the assessment of how human subjects recognize same-eye synthetically-generated iris image pairs may offer a guidance whether the fidelity of AI tools is sufficient to deliver images of non-existing subjects (thus, free from privacy issues) that could serve as training samples in human iris examiners' training programs. 

%% file: sec/2_related_work.tex
\section{Related Work}
\label{liter}

Prior work on the performance of human subjects in iris recognition has focused on ability of humans to distinguish genuine from impostor pairs of authentic irises (\ie, not AI-generated) under varying conditions. Hollingsworth \textit{et al.}~\cite{hollingsworth2011genetically} explored human evaluation of iris pairs, including left-right irises of the same individual and irises of identical twins. They found that while automated iris biometric systems could not distinguish left–right or twin comparisons from unrelated individuals. However, human observers were able to discern the overall similarity in the iris texture for both left-right iris pairs, and for identical twins.  
Guest and Stevenage~\cite{guest2013assessment} evaluated human performance by asking participants to classify iris image pairs as same or different and rate their confidence on a seven-point scale. They found that humans perform comparably to automated systems, and that combining human decisions with algorithm outputs can further reduce false acceptance rates. McGinn \textit{et al.}~\cite{mcginn2013identity} incorporated metadata such as gender, ethnicity, and approximate age and found that novice human subject achieved over 90\% accuracy.

Moreira \textit{et al.}~\cite{moreira2019performance} assessed (1) the impact of iris variations, including algorithmically easy/difficult cases, large differences in pupil dilation, disease-affected eyes, identical twins, and post-mortem samples, and (2) the effect of an annotation step to improve decision accuracy. They found that the most challenging category of pairs was the one presenting same-eye iris images with a significant difference in pupil size. The authors showed that annotation improved accuracy, particularly for this challenging case with pupil dilation differences.  In this paper we offer a different mitigation strategy for the particular case of large pupil size difference, not requiring additional human effort, and instead using the modern image-to-image translation encoder-decoder solution.

%% file: sec/3_experiments.tex
\section{Experimental Setup}
\subsection{Experimental Methodology}

Two experiments were designed, referred to as Scenario I and Scenario II, addressing pupil size variation and the inclusion of synthetic data, respectively, into human subject-based iris image matching. Both experiments were approved by the Institutional Review Board (IRB) of the University of Notre Dame, USA, and the raw data from both experiments are offered along with this paper for reproducibility. Human subjects comprised students, staff, and faculty affiliated with the institution conducting the study, and external human subjects recruited through the Qualtrics platform. All experimental sessions were conducted online, allowing participants to complete the tasks at their own pace.

The core task across both experiments involved simultaneously presenting human with pairs of segmented iris images and asking them to determine whether the images belonged to the same eye (genuine pair) or to different eyes (impostor pair). Responses were recorded on a five-point Likert scale: \say{Same (certain)}, \say{Same (probably)}, \say{Uncertain}, \say{Different (probably)}, and \say{Different (certain)}. Human subjects were not required to provide justification or annotation for their selections.

\begin{figure*}[!htbp]
    \centering
    \includegraphics[width=0.85\linewidth]{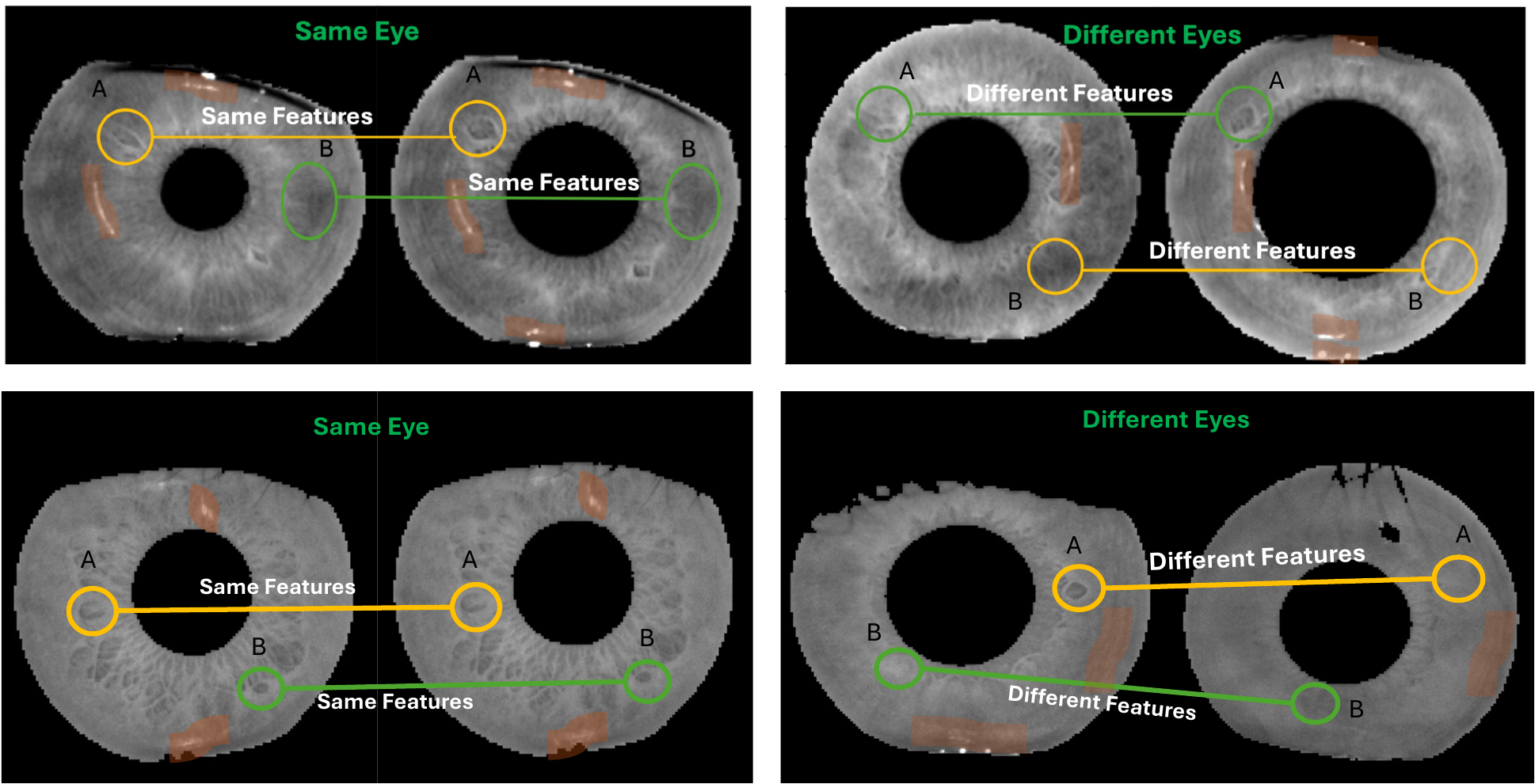}
    \caption{Sample image pairs used in a short training phase before the actual experiment. The \textbf{first row} presents an example from \textbf{Scenario I (varying pupil size)}, and the \textbf{second row} presents an example from \textbf{Scenario II (synthetically-generated samples)}. Features labeled A (yellow) and B (green) indicate iris regions relevant for decision-making, whereas the highlighted areas illustrate regions that should be disregarded during evaluation.}
    \label{fig:IIS_PDM_intro_samples}
\end{figure*}

Before starting the experiment, the instructional phase utilized two illustrative image pairs to familiarize the subjects with the task: one designated as a same-eye case and one as a different-eye case, Figure~\ref{fig:IIS_PDM_intro_samples}. To aid decision-making, key iris regions in the example pairs were annotated, and brief descriptions of relevant features were provided to the human subjects.

During the experiment, an iris pair was displayed for six seconds before the response options appeared, ensuring sufficient viewing time. Human subjects could not skip questions or submit a response immediately after displaying the images. Following response submission, a \say{Next} button became available, enabling them to advance to the subsequent image pair. To mitigate fatigue, a 20-second scheduled break was enforced after the completion of every 20 pairs. It should be noted that the two experiments were conducted independently, and thus, the human subjects group did not necessarily overlap.

\subsection{Scenario I (Varying Pupil Size)} 
\label{Exp1-PD}

This scenario was designed to (a) evaluate the performance of human subjects in discriminating between same-eye and different-eye image pairs under conditions of varying pupil size, and (b) re-evaluate this performance after aligning the pupil size. The research questions associated with these two evaluations are the following:
\begin{description}[leftmargin=1cm]
    \item[RQ1:] Should iris images be deformed prior to examination to align the pupil size in both samples, and if so, what is the best pupil size to which images should be aligned?
    \item[RQ2:] Do the deformation models (linear or non-linear) used to deform the iris texture when aligning the pupil size impact the human performance when matching aligned iris samples?
\end{description}

The alignment process, considered in the above research questions, is non-trivial and involves deforming iris textures in a way to preserve the subject's identity, and modeling the biology-specific movement of iris muscle fibers. Our experiments utilized two distinct deformation models:

\begin{itemize}
    \item {\bf nonlinear deformation}, for which an attention-based, nested U-Net autoencoder ``EyePreserve'' was used~\cite{khan2023eyepreserve}. Given an original (undeformed) iris image and a binary mask representing the desired pupil shape, the model generates a corresponding deformed iris image whose pupil shape matches the target mask, preserving the identity represented by the deformed iris while offering a deformation that mimics the non-linear way iris muscles constrict and dilate.
    \item {\bf linear deformation}, which simply follows Daugman’s \say{rubber sheet} model~\cite{daugman2002high}. It first maps the annular iris region onto a fixed-size rectangular representation, and then re-maps that rectangular region linearly onto a target annulus. 
\end{itemize}

\begin{figure*}[!htbp]
    \centering
    \includegraphics[width=\linewidth]{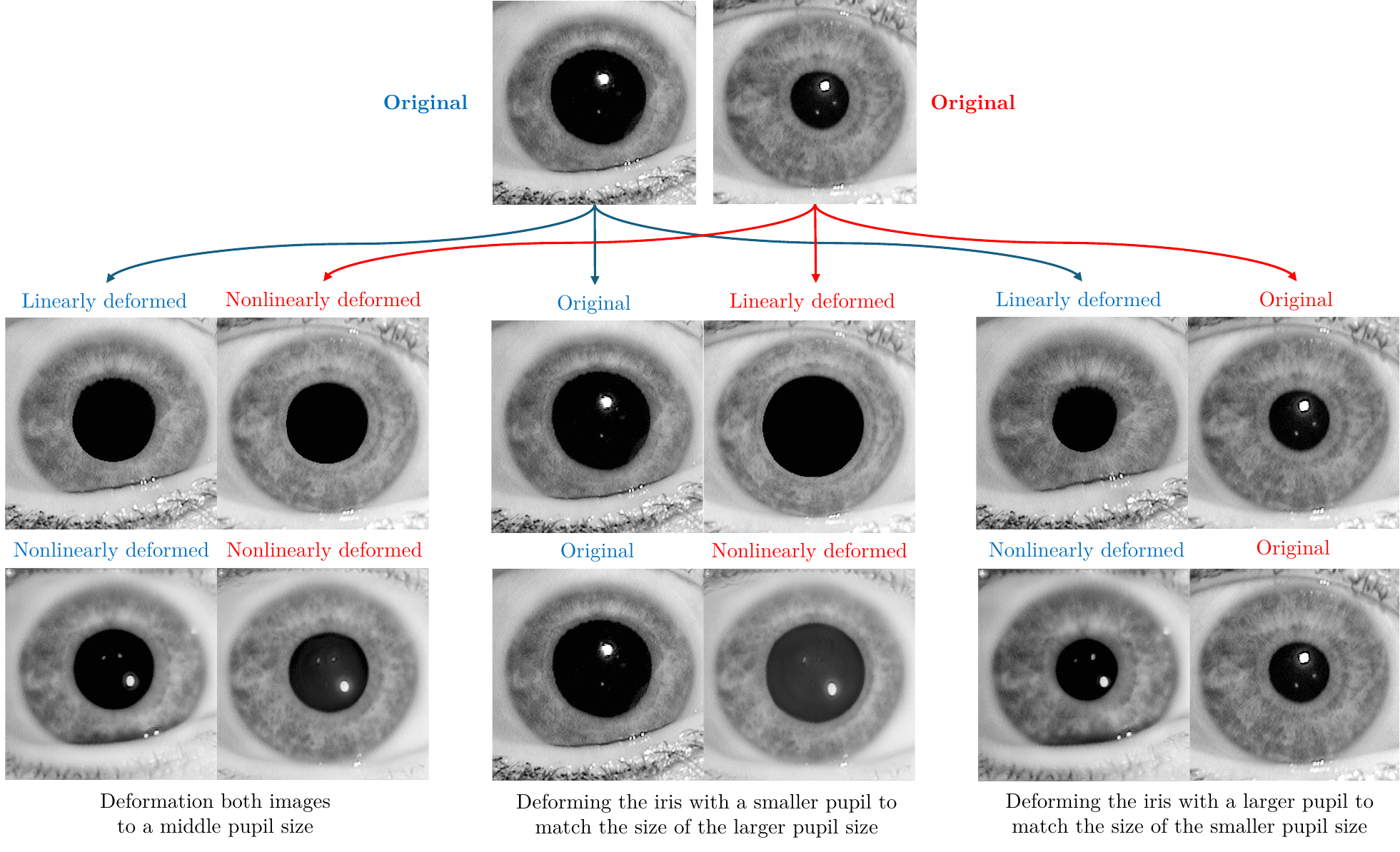}
    \caption{Seven iris texture deformation variants for one iris image pair (genuine in this example) presented to human subjects.}
    \label{fig:pupil_dynamics_Scenario}
\end{figure*}

Besides the selection of the model (linear or non-linear), an independent question relates to which image needs to be deformed. We considered the following options: (a) smaller-pupil image deformed to the larger-pupil image, (b) larger-pupil image deformed to the smaller-pupil image, and (c) both images deformed to a middle pupil size. Option (a) results in the largest degradation of the information, since the iris texture is ``squeezed'' to the smaller area. However, in this option the algorithm does not need to interpolate pixel values, hence lower chances for the so called ``hallucinations,'' at the same time making the pair harder for a human to compare. Option (b) stretches the smaller iris making it potentially easier for a human to compare, but the chances of ``hallucinating'' new information is higher. Option (c) balances between the amount of information that needs to be interpolated and the difficulty for a human to perform the matching. After adding the baseline case (no deformation applied to any of the images) and accounting for the two deformation models (linear and non-linear), we ended up with seven possible variants of each single genuine pair, all of which are illustrated in Figure ~\ref{fig:pupil_dynamics_Scenario}.
\subsection{Scenario II (Synthetically-Generated Samples)}
\label{Exp1-AS}
The goal of the second scenario is to investigate the ability of human subjects to accurately identify whether pairs of iris images originate from the same eye or different eyes, when both authentic and synthetic samples are presented. The primary objective, and thus our {\bf research question RQ3}, is to determine whether the verification accuracy achieved with synthesized same-identity (genuine) and different-identity (impostor) iris images is comparable to that obtained using authentic iris samples. 

Human subjects were simultaneously presented with image pairs drawn from four sets composed of the following image pairs: (a) authentic-genuine, (b) authentic-impostor, (c) synthetic-genuine, and (d) synthetic-impostor. These four categories were evenly distributed across the task. For each category, we included $10$ image pairs from the left eye and $10$ from the right eye, ensuring a balanced representation across eye sides and identity types. 

%% file: sec/4_dataset.tex
\section{Datasets}

\subsection{Data Source}
\label{data:source}
Authentic samples for both scenarios were sourced from the publicly available iris image datasets offered by the University of Notre Dame \cite{cvrl_datasets}\footnote{the exact filenames and dataset names are provided along with other metadata in the paper's GitHub repository: \url{https://github.com/CVRL/Human-Iris-Judge}}. To establish visual consistency and facilitate fair comparison across all presented pairs, gamma correction was applied to each image to normalize image brightness.

\subsection{Scenario I Data Preparation} We made sure that images previously utilized for training the nonlinear iris deformation model were excluded from this experiment. The selected samples exhibited various levels of pupil size by calculating pupil-to-iris ratio as defined by the ISO/IEC 19794-6 iris image quality standard \cite{iso2011iris}. faculties. 

To validate the correctness of the selected samples (genuine/impostor pair), we used the open-source HDBIF iris matcher \cite{czajka2019domain,open_source_iris}, which classifies pairs with matching scores below $0.35$ as genuine and those with scores above $0.35$ as impostor. After composing image pairs, both linear and non-linear deformation models were used to generate irises with \say{small}, \say{middle}, and \say{large} pupil sizes. In total $140$ image pairs were prepared across the four variants described in Sec.~\ref{Exp1-PD}. The reader can use Figure ~\ref{fig:Pupil_dynamic_low_high_accuracy} to see the example pairs selected for Scenario I. The Supplementary Materials include the corresponding score distributions, which illustrate the degree of separation between selected genuine and impostor samples. 

\subsection{Scenario II Data Preparation} 

For the second scenario, we trained both StyleGAN~\cite{karras2020training} and a diffusion model~\cite{nichol2021improved} on a subset of the dataset described in Sec. \ref{data:source}. Then, we generated 5,000 samples from each trained model. As we needed both authentic and synthetic samples for this experiment, we again used the HDBIF matcher to compute similarity scores between pairs of synthetic samples to select samey-eye pairs (those with a similarity score below $0.35$) and different-eye pairs (those with a score above $0.35$).
Same-eye image pairs synthesized by the StyleGAN model yielded significantly lower comparison scores than those from the diffusion model (see Supplementary Materials for the score distributions). This result indicates that StyleGAN was able to generate same-eye images that resulted in smaller distance (in the biometric feature space), thus we decided to use only StyleGAN's synthetic samples and only those that satisfied ISO/IEC 19794-6 iris image quality requirements.
 
A total of $80$ image pairs were prepared across the eight variants as described in Sec~\ref{Exp1-AS}. That is, this final set comprised $40$ pairs belonging to unique identities sourced from authentic iris samples and $40$ pairs from unique identities sourced from synthetic iris samples, ensuring an equal representation of identity sources. The reader can use Figure \ref{fig:IIS_Difficults_samples} to see example pairs in each of the eight categories.

%% file: sec/5_results.tex
\section{Results}

\subsection{Evaluation Approach}
To answer the research questions in both scenarios, we compared human judgment accuracy across different categories of iris image pairs. The $\chi^2$ test of independence was employed to determine the statistical significance of the observed differences in accuracy between categories. For all statistical analyses, the null hypothesis ($H_0$) was that there are no differences between the two distributions of human judgments being compared. Conversely, the alternative hypothesis ($H_1$) was that such a statistically significant difference does exist. A significance level of $\alpha$ set at $0.05$ was used for all tests.

\subsection{Scenario I (Varying Pupil Size)} 

A total of $100$ human subjects were invited to participate in this experiment. To answer {\bf RQ1}, we compared human judgments obtained for deformations made by the linear and non-linear models. To answer {\bf RQ2}, we compared human judgments made for (a) original (non-deformed) iris images, with those in which (1) one of the images was deformed to match the larger pupil size, (2) one of the images was deformed to match the smaller pupil size, and (3) both images were deformed to a middle pupil size (corresponding the average size of the pupil calculated from two iris images being compared).

\begin{table*}[!htbp]
\centering
\caption{Accuracy of human subjects' judgments in Scenario I observed for various deformation approaches, showed for genuine, impostor, and combined image pairs. \textit{Large} means that the iris with a smaller pupil size is deformed to match the iris with a larger pupil size. \textit{Small} means that the iris with a larger pupil size is deformed to match the iris with a smaller pupil size. \textit{Middle} means that both images are deformed to match a canonical, middle-sized pupil. The $p$-value $<0.001$ in each $\chi^2$ test comparing human judgments obtained for different groups of pupil size (small vs middle, small vs large, and middle vs large), what suggests that the observed differences are statistically significant at the assumed statistical significance level $\alpha = 0.05$.}
\label{tab:number_of_responses_PD}
\begin{tabular}{c|c|c|c|c|c}
\toprule
{\bf Deformation} & {\bf Pupil} &
\multicolumn{4}{c}{\bf Human Subject Accuracy (\%)} \\
\textbf{Method} & \textbf{Size} & Genuine Pairs & Impostor Pairs & All Pairs (Combined) & Per Deformation Model\\
\midrule
None & Original & $75.80$ & $77.00$ & $76.40$ & --- \\
\hline
\multirow{3}{*}{Nonlinear} 
 & Large  & $76.40$ & $68.40$ & $72.40$  \\
 & Middle & $\bf90.30$ & $76.50$ & $\bf83.40$ & $77.17$\\
 & Small  & $74.8$0 & $76.60$ & $75.70$ \\
\hline
\multirow{3}{*}{Linear} 
 & Large  & $79.40$ & $68.50$ & $73.95$ \\
 & Middle & $85.30 $& $79.80$ & $82.55$ & $\bf79.67$ \\
 & Small  & $83.70$ & $\bf81.30$ & $82.50$ \\
\bottomrule
\end{tabular}
\end{table*}

Table~\ref{tab:number_of_responses_PD} summarizes the human subject accuracies obtained across different variants, including original iris image pairs and pairs of deformed images. The following observations are noted from a comparative analysis of recognition accuracies in this table:

\begin{enumerate}[label=\alph*),leftmargin=4mm]

   \item For the original genuine iris pairs without any pupil shape manipulation, human subjects exhibited lower accuracy compared to the pairs, in which iris were deformed to match pupil sizes. The only exception to this trend was when the iris image with the larger pupil was deformed to match the one with the smaller pupil using the nonlinear deformation model. The latter may be caused by the fact that the model, when significantly stretching the iris texture, had to ``interpolate'' information that was missing due to biology of iris muscle movement (\eg, folding the pupillary frill for dilated pupil).
   
   \item Deforming genuine pairs to a middle pupil size resulted in the highest overall accuracy for both deformation methods (linear and nonlinear). Additionally, humans achieved approximately five percentage points higher accuracy when the images were deformed with the nonlinear model, compared to linear deformations.
   
   \item For impostor pairs, nonlinear deformation did not improve human accuracy ($77.0$\% for original vs. highest accuracy of $76.6$\% for deformed images). However, deforming to a small pupil size using the linear model slightly increased the accuracy (from $77.0$\% to $81.3$\%).
   
   \item For both genuine and impostor comparisons, deformation to a middle pupil size yielded slightly higher accuracy with the nonlinear model compared to the linear model ($83.4$\% vs. $82.55$\%, respectively) but this difference was not statistically significant.
   
   \item As expected, deforming images to a large pupil size reduced recognition accuracy for both genuine and impostor pairs in all deformation methods, likely due to the loss of fine-grained iris texture cues essential for accurate comparison.
\end{enumerate}

In general, deforming iris images to a middle pupil size resulted in the highest overall accuracy ($82.98$\%), whereas pairs deformed to a large pupil size exhibited the lowest accuracy ($73.18$\%). These observations offer the direct answer to our {\bf RQ1} posed in Sec. \ref{Exp1-PD}.

Interestingly, human subjects achieved slightly higher total accuracy ($79.67$\%) when the images were deformed linearly, compared to a nonlinear deformation ($77.17$\%), and the Chi-squared test suggests that this difference is statistically significant ($p$-value $< 0.001$). Hence, when images are not aligned to a middle pupil size, linear deformation is sufficient. This finding answers our research question {\bf RQ2} posed in Sec. \ref{Exp1-PD}. 

\begin{figure*}[!htbp]
    \centering
\includegraphics[width=0.85\linewidth]{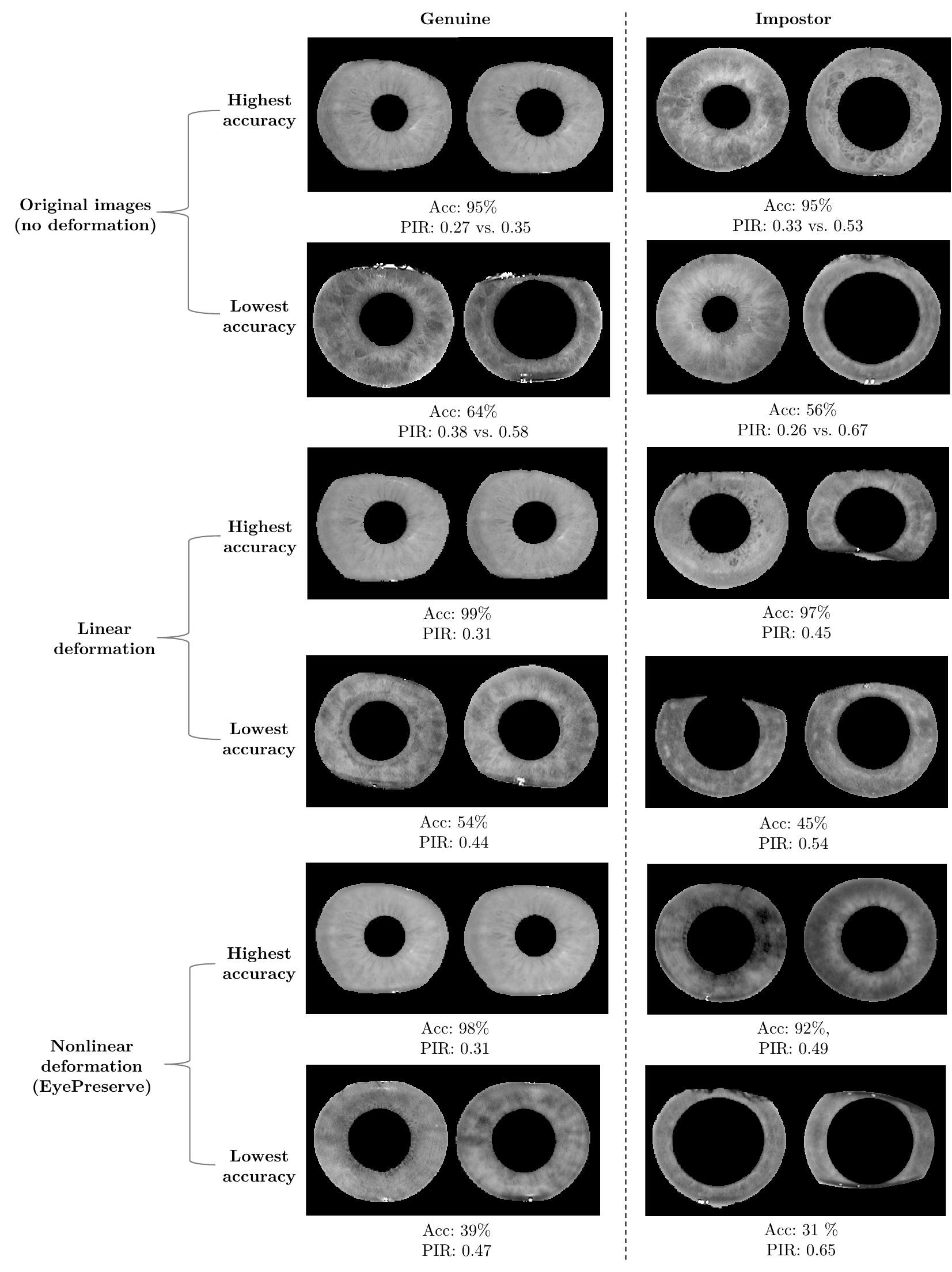}
    \caption{Image pairs from Scenario I with the lowest and highest human subjects’ accuracy for both non-deformed and deformed samples. ``Acc'' denotes accuracy, and ``PIR'' represents the pupil-to-iris ratio. The reported accuracy value corresponds to the average accuracy computed over all $100$ human subjects.}
\label{fig:Pupil_dynamic_low_high_accuracy}
\end{figure*}

Fig.~\ref{fig:Pupil_dynamic_low_high_accuracy} presents the image pairs with the lowest and highest human accuracy, separated by pair type (genuine or impostor) and deformation variant (none, linear, or non-linear). For image pairs with identical pupil sizes, a single pupil-to-iris ratio (PIR) value is reported. Otherwise, two distinct values are shown for the left and right images. This Figure illustrates that iris samples with very large pupil sizes were associated with notably lower recognition accuracy and present a significant difficulty for the subjects to make a correct determination for iris images with highly dilated pupils.


\subsection{Scenario II (Synthetically-Generated Samples)} 

Another $163$ human subjects were invited to participate in this experiment. We compared human judgments made for same-identity and different-identity image pairs representing both the authentic and synthetic irises.

\begin{figure*}[!htbp]
    \centering
    \begin{subfigure}[b]{0.49\textwidth}
        \centering
        \includegraphics[width=1.0\textwidth]{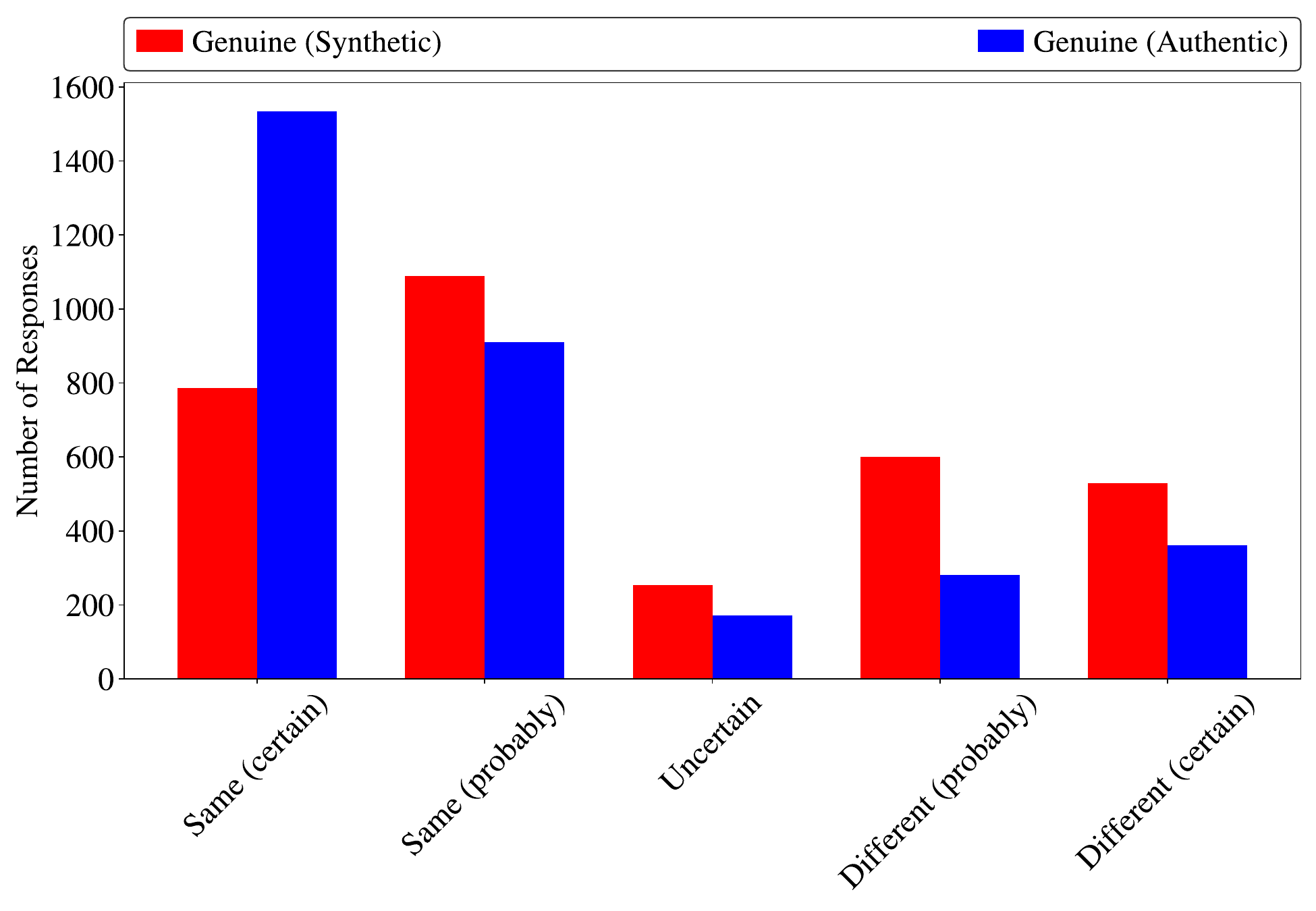}
        \caption{}
        \label{fig:synthetic}
    \end{subfigure}
    \hfill
    \begin{subfigure}[b]{0.49\textwidth}
        \centering    
        \includegraphics[width=1.0\textwidth]{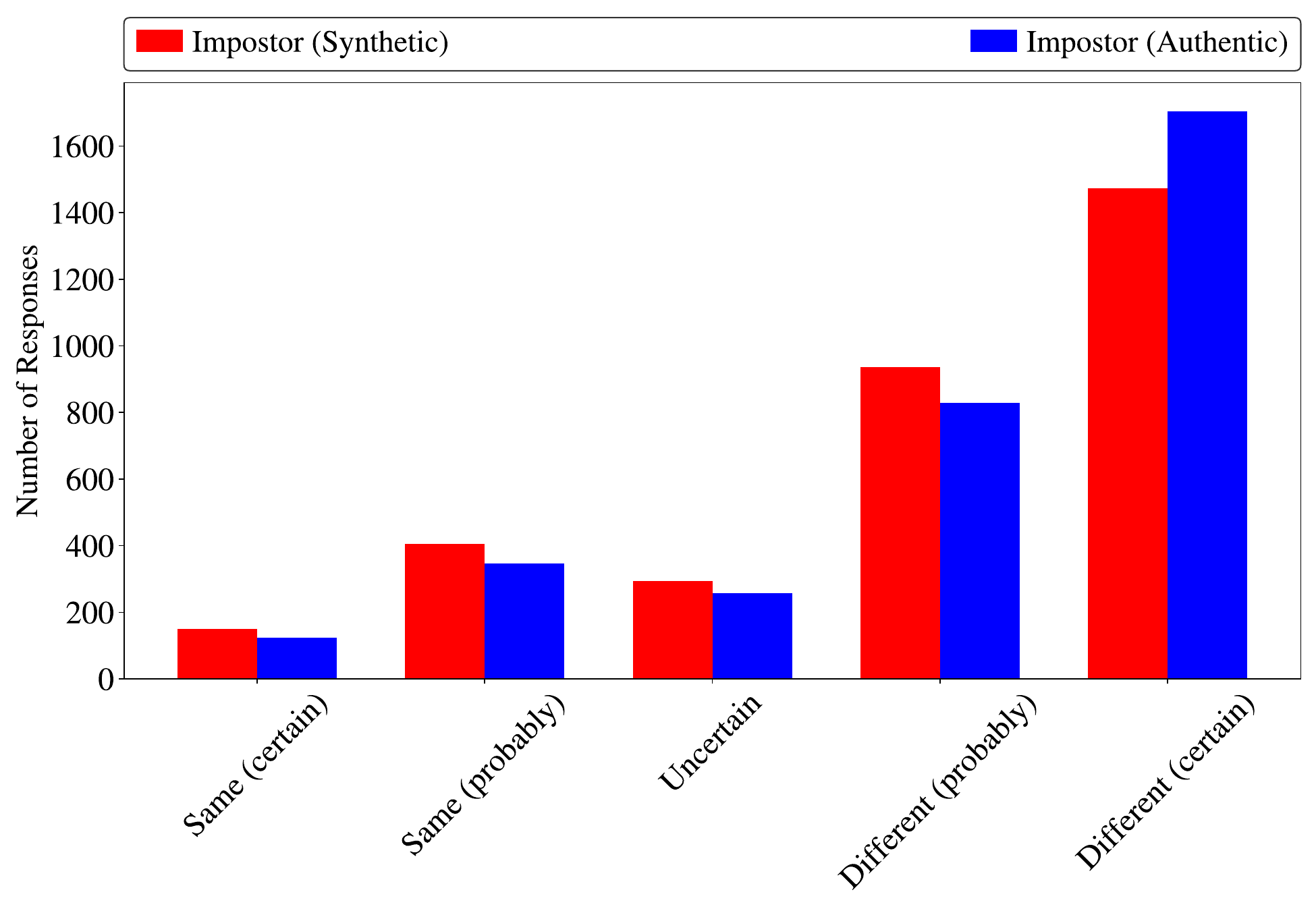}
        \caption{}
        \label{fig:imposter}
    \end{subfigure}
    
    \caption{The number of human subject responses for (a) genuine (authentic vs. synthetic) and (b) impostor (authentic vs. synthetic)
     iris pairs in Scenario II as a function of confidence level of the subjects.}
    \label{fig:IIS_Response_Plot}
\end{figure*}

\begin{table}[!htbp]
\caption{Accuracy of human subjects in identifying genuine and impostor iris pairs in Scenario II for authentic and synthetic image pairs showed separately.}
\label{tab:IIS_Response}
\centering
\begin{tabular}{c|c|c}
\toprule
{\bf Pair} &
\textbf{Iris Sample Type} &
\textbf{Accuracy (\%)} \\
\midrule
Genuine & Authentic Only & $75.00$ \\
 & Synthetic Only & $57.55$ \\
\hline
Impostor & Authentic Only & $77.73$ \\
 & Synthetic Only & $73.93$\\
\bottomrule
\end{tabular}
\end{table}

Figure~\ref{fig:IIS_Response_Plot} and Table~\ref{tab:IIS_Response} summarize the counts of correct and incorrect human responses by type of image pair (authentic or synthetic). As a reminder, the question was whether an iris image pair represents the same eye or different eyes. We can see that human subjects were more accurate in identifying impostor pairs than genuine ones. Moreover, human accuracy was higher for authentic samples, particularly for genuine–authentic pairs compared to genuine–synthetic pairs. 

Conducting $\chi^2$ test between different group of genuine/impostor and authentic/synthetic samples indicates that the differences among the variants presented in Table~\ref{tab:IIS_Response} are statistically significant. The corresponding $p$-value $<0.0001$ for all genuine–authentic vs. genuine–synthetic, impostor–authentic vs. impostor–synthetic, and genuine vs. impostor pairs.

Although the $\chi^2$ test confirmed a statistically significant difference between impostor–authentic and impostor–synthetic accuracies, the magnitude of this difference is low ($77.73$\% vs. $73.93$\%). This result provides an empirical response to the research question \textbf{RQ3}, indicating that human subjects exhibit slightly better performance when assessing authentic iris samples compared to StyleGAN-generated synthetic counterparts.

\begin{figure*}[!htbp]
    \centering  \includegraphics[width=\linewidth]{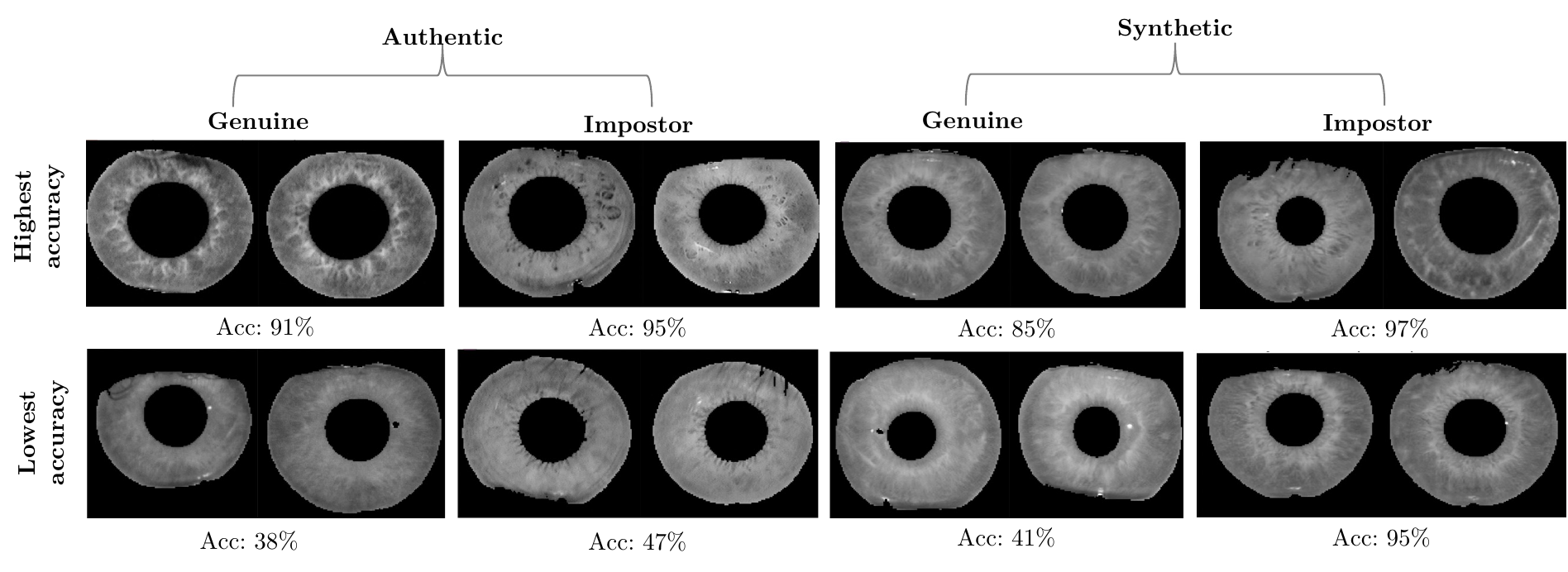}
    \caption{Image pairs from Scenario II with the lowest and highest human subjects’ accuracy. Here ``Acc'' stands for accuracy and represents the average performance accuracy of all $163$ human subjects.}
    \label{fig:IIS_Difficults_samples}
\end{figure*}

Figure~\ref{fig:IIS_Difficults_samples} presents the image pairs with the lowest and highest human accuracy, separated by pair type (genuine or impostor) and sample type (authentic or synthetic). As anticipated, pairs displaying more pronounced similarities or differences were classified with greater accuracy, highlighting the role of salient visual cues in human decision-making. In contrast, pairs exhibiting subtle iris textures or minimal feature correspondence were associated with the lowest accuracy, suggesting that the absence of distinctive features substantially increases classification difficulty.

%% file: sec/6_conclusion.tex
\section{Conclusions}
This study, for the first time according to the authors' knowledge, examined human performance in iris verification under two controlled scenarios: (a) varying pupil size with and without pupil size alignment, and (b) the use of synthetically-generated samples. We hired $263$ human subjects in total to compare pairs of iris images and judge whether they represent the same eye or different eyes. 

In \textit{Scenario I}, we assessed the performance of human subjects in matching iris images with identical or different pupil sizes. The results indicate that pupil size normalization substantially enhances recognition accuracy, with alignment to a medium-sized pupil yielding the highest performance. Larger pupil deformations reduced accuracy, likely due to decreased iris texture visibility and fewer iris texture features available for comparison. Although the linear deformation model slightly outperformed the nonlinear model, the nonlinear approach was more effective when both images in a pair were aligned to the middle pupil size.

In \textit{Scenario II}, we assessed the performance of human subjects in matching iris images representing authentic and synthetically-generated same-eye and different-eye irises. The results showed a higher accuracy of human subjects in distinguishing same-eye and different-eye pairs for authentic samples compared to synthetic ones, particularly for genuine pairs. For impostor pairs, authentic samples also led to slightly higher accuracy. All observed differences were statistically significant. This may suggest that the fidelity of same-eye synthetic iris images, offered by contemporary generative models, has not yet reached the fidelity of authentic same-eye iris images.

The human judgments from both experiments, along with pointers to the files sourced from public iris image datasets, as well as segmentation masks are offered along with the paper\footnote{\url{https://github.com/CVRL/Human-Iris-Judge}} for replicability purposes.